# Morphological Detection and Classification of Microplastics and Nanoplastics Emerged from Consumer Products by Deep Learning

Hadi Rezvani[1 ‡], Navid Zarrabi[2 ‡], Ishaan Mehta[2], Christopher Kolios[3], Hussein Ali Jaafar[2], Cheng-Hao Kao[2], Sajad Saeedi[4]*, and Nariman Yousefi[1]*

*‡ Authors contributed equally to the paper.*

[1] Department of Chemical Engineering, Toronto Metropolitan University, Toronto, Canada

[2] Department of Mechanical, Industrial, and Mechatronics Engineering, Toronto Metropolitan University, Toronto, Canada

[3] Department of Computer Science, Toronto Metropolitan University, Toronto, Canada

[4] Department of Computer Science, University College London, London, England

*Corresponding Authors:

Sajad Saeedi – Email: s.saeedi@ucl.ac.uk

Nariman Yousefi – Email: nariman.yousefi@torontomu.ca

**Abstract**

Plastic pollution presents an escalating global issue, impacting health and environmental systems, with microplastics and nanoplastics found across various compartments, including potable water and air. Manual methods for studying these contaminants are labour-intensive and time-consuming, necessitating a shift towards more automated techniques. We present here the Micro- and Nanoplastics Dataset (MiNa), a novel and open-source scanning electron microscopy (SEM) micrograph collection engineered for automatic detection, quantification, and segmentation of MNP particles with morphologies mimicking MNPs in real aquatic environments. The dataset, comprising SEM images obtained from particulate plastics that emerged under realistic conditions, categorizes them by polymer type across a broad size spectrum. These micrographs were trained using an AI-enhanced pipeline for annotation and were subsequently used to train various deep learning models for automatic identification of MNPs in any given SEM micrographs. We demonstrate the application of commonly used algorithms in the literature on MiNa dataset, evaluate their effectiveness, and identify the unique challenges and potential of each method. We also explore the limitations of existing generalist foundation models for micrographs and the potential for using fine-tuned models to provide domain-specific prompts. Our results show that deep learning methods can achieve reliable automated detection, segmentation, and classification of MNPs, while also highlighting key limitations related to particle scale, crowding, and domain-specific morphology. We further propose a domain-specific prompting strategy in which fine-tuned models guide foundation models, improving the generalization and robustness of MNP segmentation in complex micrographs, and validate its performance on external microplastic datasets to assess scalability.

**Synopsis**

We report an open-source and fully annotated scanning electron micrograph dataset of microplastics and nanoplastics, which was used for successfully training deep learning algorithms for automatic detection, quantification, and segmentation of particulate plastics.

## Visual Abstract

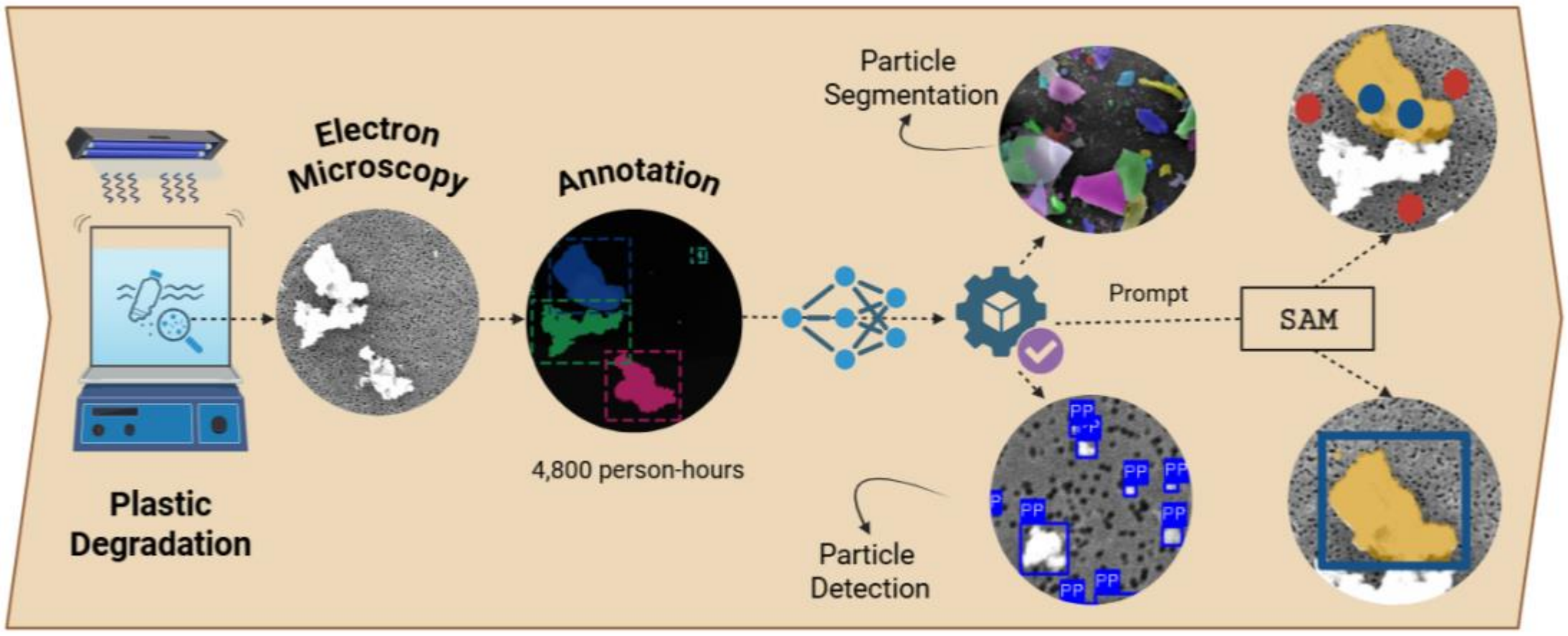

## 1. Introduction

Plastic pollution in aquatic environments, especially in the Great Lakes where around 22 million pounds of plastic enter annually[1], poses a significant environmental challenge. The breakdown of consumer plastics into microplastics (less than 5 $mm$) and nanoplastics[2] (less than 1 $\mu m^2$) increases their surface area and mobility, allowing them to absorb and carry harmful substances like persistent organic pollutants (POPs)[3] and heavy metals[4]. This makes microplastics vectors for pollutants, aiding their movement through the food chain. While conventional water and wastewater treatment plants effectively remove larger plastic particles, smaller micro- and nanoplastics (MNPs) in drinking water remain a concern[5]. For example, MNPs have been long detected in our food chain[6]. Furthermore, intergenerational transfer of nanoplastics has been observed, indicating their potential mutagenic and carcinogenic properties[7]. This is primarily attributed to the plastic additives and other contaminants that MNPs carry, highlighting the need for effective measures to address their presence in the environment and food chain.

Traditional MNP analysis, including Raman and Fourier transform infrared (FTIR) spectroscopies, pyrolysis-gas chromatography, and various microscopy techniques, remains limited by manual sampling and labor-intensive processing[8–10]. MNP counting relies mainly on manual analysis of microscopy images, with image processing software aiding the analysis. These manual workflows are prone to error, high cost, and are insufficient for large-scale environmental monitoring[11,12], necessitating automated, high-throughput approaches for MNP detection and quantification.

Quantification of MNPs in environmental samples is challenging due to particle aggregation, biological fouling, and low environmental concentrations[13,14]. Furthermore, traditional microscopy and spectroscopy require complex, time-consuming sample preparation, limiting high-throughput analysis[15]. Beyond physical complexities, human-centric visual identification remains unreliable, with success rates as low as 39–68%[16]. Consequently, researchers are transitioning toward automated machine learning (ML) pipelines, initially trained on simplified datasets of synthetic latex beads[17], ball-milled fragments[18], or processed environmental samples[19–21].

While ML enhances traditional identification techniques, each analytical method presents specific constraints. FTIR and Raman spectroscopies provide chemical insights but are limited in analyzing nanoplastic size limits, fluorescence interference, and laborious pretreatment[22,23]. Conversely, optical and electron microscopy provide the high-resolution morphological data required for precise size and shape analysis of complex, weathered particles[24]. Recently, deep learning methods were proven effective in learning detailed morphological features in microscopy images.

Several deep learning approaches have been used for learning morphological features in micrographs (Table 1). Detection of particles in micrographs is possible with region-based detectors for bounding box extraction[25] or semantic segmentation requiring post-processing to isolate individuals[12,26]. Conversely, instance segmentation frameworks like Mask R-CNN provide a unified architecture for simultaneous localization and pixel-level segmentation without additional processing steps[27]. In addition, YOLO algorithms[28] offers an efficient single-stage alternative to two-stage R-CNN architectures, maintaining accuracy in complex scanning electron microscopy (SEM) images with overlapping or variably sized particles[29]. More recently, transformer-based architectures and foundation models have emerged for microscopy[30]**;** however, their efficacy for identifying MNPs is limited by the irregular morphologies of real-world environmental debris. The lack of a foundation model optimized for weathered MNPs underscores the need for large-scale, realistic annotated datasets. Such data would also support the development of self-supervised transformer pipelines capable of leveraging unlabeled SEM imagery[31].

This work utilizes accelerated weathering of consumer plastics (e.g., bottles, bags, packaging) to generate environmentally relevant MNPs (<1 μm). We introduce the Micro- and Nanoplastics (MiNa) dataset, the first open-source SEM collection of weathered MNPs annotated by polymer type, such as polyethylene (PE), expanded polystyrene (PS), polypropylene (PP) and polyethylene terephthalate (PET), rather than just physical shape. This enables a systematic investigation of morphology-chemistry relationships across a broader size range than previously available[12,26]. Using MiNa, we benchmark representative deep learning architectures, including Faster R-CNN, YOLO, RT-DETR, and Mask R-CNN, and evaluate the zero-shot capabilities of foundation models through domain-specific prompting (Figure 1**)**. By providing the first publicly accessible, instance-level annotated dataset for degraded consumer polymers[32], this work addresses the lack of open data[25–27] and supports reproducible automated MNP monitoring.

**Table 1.** Datasets for automatic detection of MNPs based on images using deep learning

| Source | Benchmarking Algorithms | Labels | MNP Size | Public |
|---|---|---|---|---|
| Digital Camera Imaging (2021)[26] | Particle Segmentation (Unet + VGG16) | Fragments, Pellets, Lines | 1 - 5 mm | No |
| Optical Cameras Imaging (2023)[27] | Particle Segmentation & Shape Classification (Mask R-CNN) | Fiber, Fragment, Pellet, Rod | 0.85 - 4.76 mm | No |
| Scanning Electron Microscopy (2022)[12] | Particle Segmentation (U-net + Mean Shift), Shape Classification (VGG16) | Fiber, Pellet, Fragment | 50 µm - 1 mm | Yes |
| Fluorescent Microscopy (2023)[25] | Particle Detection & Shape Classification (Improved Faster R-CNN) | PE, PS | 5 - 20 µm | No |
| Digital Camera + TinyScope (2025)[33] | Particle Detection (YOLOv5) | Microplastics (Single class) | Not Mentioned | Yes |
| Microscope Image (2024)[34] | Particle Segmentation & Classification (U-net, EfficientNetV2B3 + MRFM) | Fragment, Fiber | 50um-10mm | Yes |
| Scanning Electron Microscopy (Our Work, 2026) | Particle Segmentation (YOLOv8-seg & v26-seg, Mask R-CNN R101& X101, SAM & Micro-SAM), Particle Classification (Faster R-CNN R101 & X101, YOLOv8 & v26, RT-DETR) | PE, PS, PP, PET | 1 µm - 1 mm | Yes |

## 2. Materials and Methods

### 2.1. Sample Preparation Process

Four of the most common types of consumer plastics were chosen for this study. Two different types of take-out food containers made from PP and expanded PS, as well as low density PE plastic shopping bags, and PET water bottles were obtained from local Toronto grocery stores and restaurants. These plastics were chosen due to their widespread application and environmental relevance. Each class represents a major category of consumer polymers with distinct physical properties, applications, and recycling challenges, highlighting the importance of their identification and classification in environmental studies and recycling efforts.

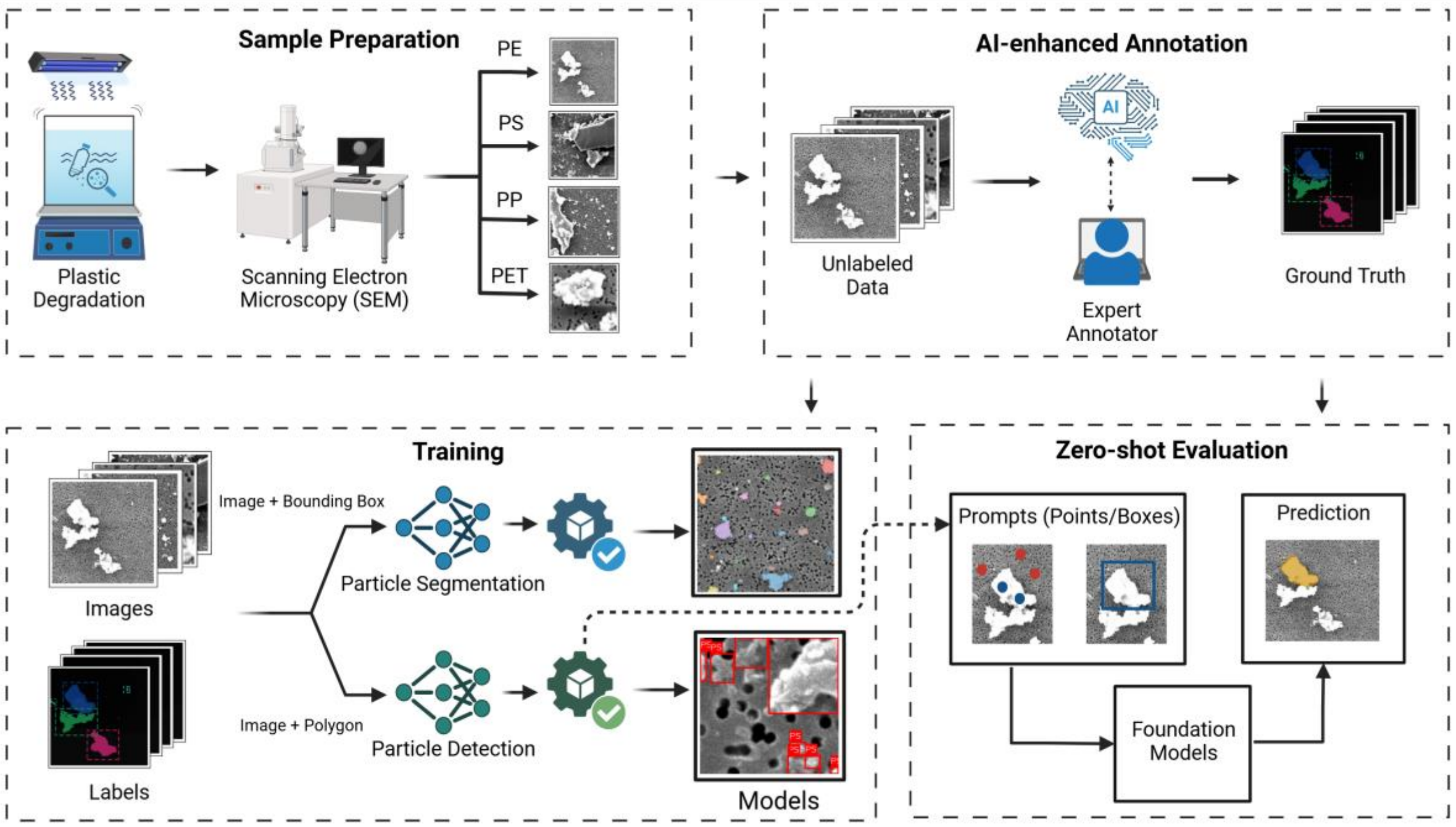

**Figure 1.** Polymer degradation under realistic simulated conditions generated fragments of MNPs from consumer products. SEM micrographs of these particles were collected and annotated through an AI-enhanced workflow, producing both bounding-box and polygon-based ground-truth labels. The resulting dataset was used to train and benchmark object detection and instance segmentation models. Furthermore, zero-shot segmentation performance was evaluated using the Segment Anything Model (SAM), including both the generalist model and fine-tuned variants, and an effective prompting strategy for zero-shot segmentation was recommended.

MNPs were generated from bulk polymers through accelerated environmental weathering, as verified in our previous work[35]. To prepare samples of MNPs, a controlled experimental setup was arranged within an enclosed chamber equipped with UV and visible light sources for accelerated degradation of the bulk plastics. UV light, with an average wavelength of 200 nm, was selected to accelerate degradation and amplify the degradation rate. Fluorescent light was used to simulate visible range wavelengths. Glass photoreactors were placed within the chamber equipped with a shaker to simulate the mechanical abrasion process. Each photoreactor contained 300 ml of deionized (DI) water. Within each photoreactor, nine pieces of each polymer (2 × 2 $cm^2$) were immersed as sources of MNP production. Control reactors containing DI water with no bulk plastics were also included to monitor potential environmental contamination by dust and other particles. A visual representation of the experimental setup is provided in Figure S1.

The morphology of the polymers before and after degradation, as well as the morphology of the emerged MNPs, were studied with a Zeiss EVO LS15 Scanning Electron Microscope (SEM) equipped with a Bruker Energy-Dispersive X-ray spectroscopy (EDS) system, operated at an acceleration voltage of 20 kV using secondary electrons. After 12 weeks of degradation, MNPs were collected by vacuum filtration of a 50 ml aliquot from the aqueous medium using track-etched polycarbonate membranes (MilliporeSigma™ Isopore™ Membrane Filters) with 200 nm pores. Subsequently, the membranes were gold-coated, and SEM images were captured. The magnification of the obtained micrographs varied from 100× to 1000×.

## 2.2. Data Organization

The MiNa dataset contains 105 SEM images, each with a resolution of 1280 × 960 pixels and featuring a scale bar at the bottom for size reference. Figure S2 shows typical SEM images of each class in the dataset. These images depict pure samples, with each one showcasing microplastics from the previously mentioned categories. To support multiple analytical tasks, the dataset was structured for both instance segmentation and object detection by providing polygon masks and bounding box annotations. All particles are labeled according to their source polymer type. Each annotation additionally includes particle area and equivalent diameter, enabling performance evaluation as well as quantitative analysis of morphology-chemistry relationships in degraded MNPs.

## 2.3. AI-assisted annotation workflow

Particles were annotated on the V7 platform through a 4,800 person-hour effort involving six annotators over five months. To manage high-density micrographs, we used a hybrid segmentation workflow, comprising (i) a custom GUI utilizing hysteresis thresholding for real-time, manual boundary setting, and (ii) point-prompts via the Segment Anything Model (SAM)[36] for zero-shot contour generation. Although SAM accelerated the process, its limitations in low-contrast and small-particle regions necessitated manual refinement. Accuracy was ensured through two rounds of cross-validation and automated post-processing to resolve overlapping contours and artifacts (e.g., "holes" or noise from accidental clicks). This multi-stage review minimized annotation bias and ensured high-fidelity segmentation in crowded image domains.

## 2.4. Data Statistics

The MiNa dataset comprises 105 micrographs across four polymer types (Tables S1). Morphological characterization included diameter, area, and key shape descriptors including aspect ratio, form factor, roundness, and convexity[37]. To assess the feasibility of morphology-based polymer identification, statistical variance was evaluated. Normality and homogeneity were tested using Shapiro-Wilk and Levene's tests, respectively. Given the non-normal distribution of morphological metrics ($p < 0.05$), the non-parametric Kruskal-Wallis test was employed to identify significant differences across polymer classes. Significant results were further analyzed via post-hoc Dunn's test with Bonferroni correction for multiple pairwise comparisons. This statistical framework ensures that observed morphological variations are mathematically distinct and representative of each degradation-induced polymer type. Details of this analysis can be found in Section S1.

## 2.5. Configuration of the Dataset

The SEM images from the MiNa dataset were split into image patches using a sliding window method. The square patches were generated in 128, 256, and 512-pixel variants. Among these, the $256 \times 256$ pixel$^2$ patches yielded the best results from the networks, providing an optimal balance between detail preservation and computational efficiency. This finding aligns with existing literature[38]. The patches were selected using a sliding window with a step size of 256 pixels to avoid overlap. Patches consisting entirely of background or particle pixels were excluded during post-processing. The remaining patches were randomly divided into 70% training, 15% validation, and 15% test splits. Additionally, annotation files are stored in both COCO[39] and YOLO[28] formats, two of the most commonly used formats in object detection.

## 2.6. Metrics

MiNa provides polygon and bounding box annotations for dual-task evaluation. Instance segmentation performance was assessed using average precision (AP) across various IoU thresholds and particle sizes and mean absolute percentage error (MAPE) quantified morphological and size prediction accuracy. Object detection was evaluated via precision, recall, and F1-score. Details of metric definition have been included in Section S2.

### 2.7. Benchmarking

We evaluated representative one-stage (YOLO, RT-DETR) and two-stage (Faster R-CNN, Mask R-CNN) architectures, widely adopted for SEM-based microplastic analysis[25,27,40]. Models were trained on a computer with Intel Core i7-13700K CPU, 16 GB RAM, NVIDIA RTX 4080 GPU, and NVME SSD (hyperparameters in Table S2). Experiments were partitioned into (i) class-agnostic instance segmentation to quantify irregular particle morphology, and (ii) polymer-specific object detection (PE, PS, PP, PET) to assess if degradation-induced morphological cues enable data-driven polymer differentiation. Finally, we benchmarked zero-shot foundation models, utilizing fine-tuned detectors as domain-specific prompting engines to improve MNP segmentation. This workflow established a baseline for automated MNP identification and investigated the relationship between polymer chemistry and weathered morphology in complex image domains.

## 3. Results and Discussions

### 3.1. Characteristics of the Dataset

The data extracted from the MNP SEM micrographs derived from various bulk polymers is invaluable for understanding plastic pollution and making informed environmental decisions. Figure 2 illustrates the MNP particle size distribution in different classes of polymers. PS released 585 nanoplastics (defined as particles smaller than 1 μm), while PP had a significantly higher count of 3,078 nanoplastics. PET and PE had 2,261 and 153 particles in the nanoscale, respectively. The high number of nanoplastics released from PP and PET highlights the potential for these polymers to contribute to nanoscale plastic pollution, which can impact biological tissues and ecosystems more significantly in comparison with microplastics. The analysis revealed that PP and PS exhibit the highest fragmentation into microplastics in the 0-5 μm range. Specifically, PP showed a significant number of particles (6,568), while PS followed with 4,744 particles. This suggests that under the same photodegradation regime, PP and PS were more susceptible to breaking down into smaller particles compared to PET and PE, which showed lower counts of 3,486 and 513 particles, respectively. The high fragmentation rate of PP and PS into microplastics highlight their potential environmental impact, as smaller particles can disperse more widely and pose greater risks to marine and terrestrial ecosystems. In the 5–50 µm range, PS displayed the highest number of particles, totaling 6,412. This is substantially higher than PP (1,383), PET (438), and PE (100).

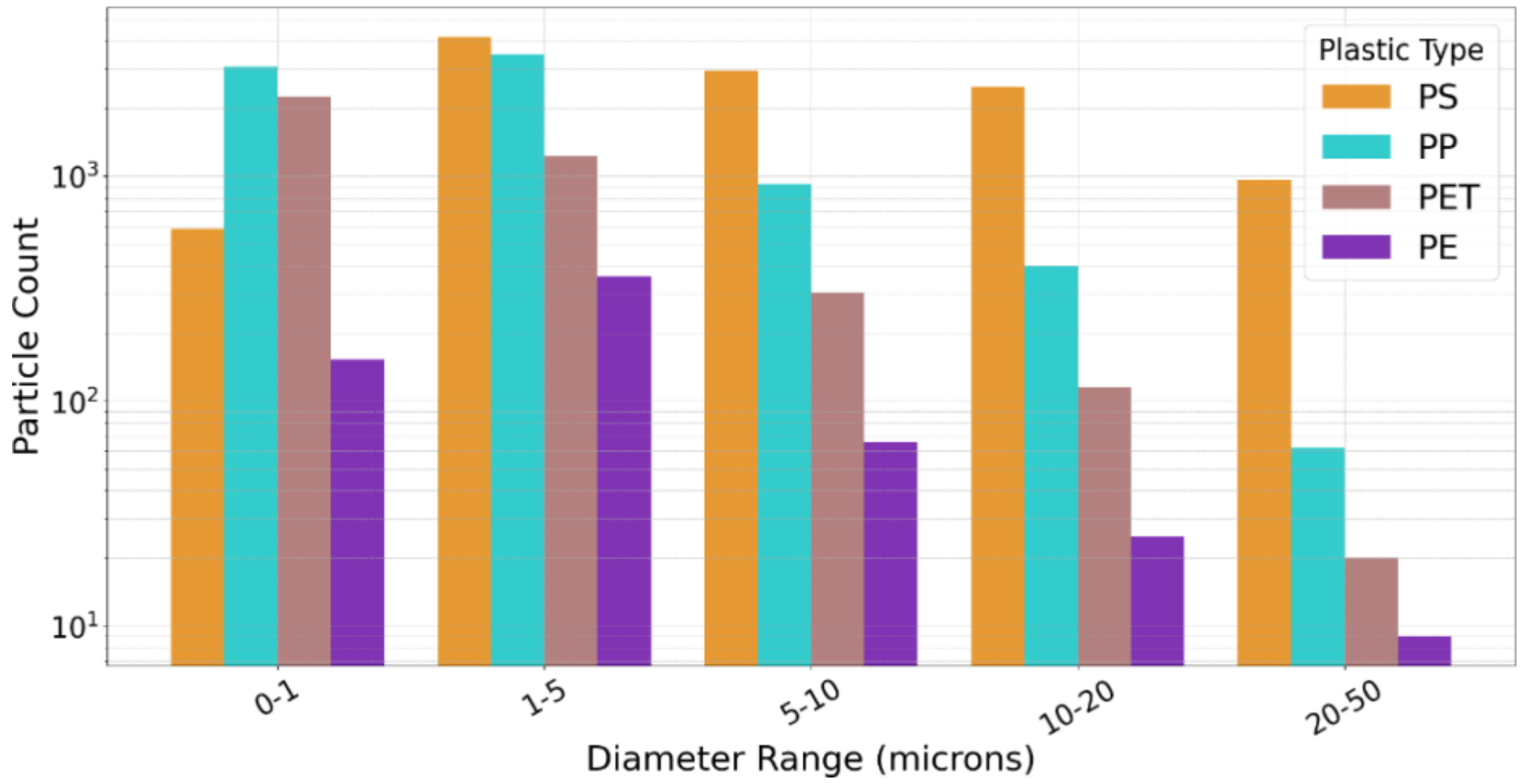

**Figure 2.** Size distribution of the MNPs emerged from bulk PS, PP, PET, and PE in the course of 12 weeks of photodegradation.

The data indicate that PS is prone to fragmenting into this size range more extensively than the other polymers. The relatively lower counts for PP, PET, and PE in this range suggest a less frequent occurrence of these polymers breaking into larger microplastic particles. This could be attributed to differences in their chemical structure and resistance to environmental degradation.

Larger particles (50–100 µm) are less prevalent across all polymer types, with PS having the highest count at 99 particles. PP and PE exhibit minimal counts (5 and 1, respectively), while PET has no particles in this range. The scarcity of large particles indicates that these fragments likely have further broken down into smaller sizes over time. The higher count for PS may suggest an initial stage of fragmentation before further degradation into microplastics. Based on Table 2, PS exhibited the highest number of emerged particles (11,272 particles) with an average diameter of 9.352 µm and a significantly large average area of 50.032 $\mu m^2$. In contrast, PE showed the lowest particle count (614 particles) but a larger average diameter of 3.175 µm and an average area of 8.313 $\mu m^2$ compared to PP and PET. These differences in metrics highlight the distinct morphological characteristics of particles in each polymer type, which are crucial for understanding their behavior and applications in various fields. The data on aspect ratio, form factor, roundness, and convexity further provide insights into the shape and structural properties of the particles, contributing to a deeper understanding of the degradation of bulk polymers into MNPs. Based on the Dunn's test with Bonferroni correction for diameter, significant differences were found between all pairs except PET and PE. Similarly for area, all pairs showed significant differences apart from PET and PE. Aspect ratio showed significant differences between PS and PP, PP and PET, and PP and PE, but no significant differences between PS and PET, PS and PE, and PET and PE. Form factor showed significant differences between all pairs except PP and PET. For roundness, significant differences were found between all pairs except PET and PE. Convexity showed significant differences between all pairs except PP and PET. The statistical analysis confirmed that the morphological characteristics of microplastics varied significantly between the different polymer types. These differences are crucial for understanding the environmental impact and behavior of microplastics. For example, MNPs can act as vectors for dissolved contaminants which is driven by their exposed surface area. The significant variation in properties like diameter, area, form factor, and roundness across polymer types highlights the need for tailored approaches in microplastic management and mitigation strategies.

**Table 2.** Comparison of particle metrics (average) across different bulk polymers types

| | **No. of Particles** | **Diameter (µm)** | **Area ($\mu m^2$)** | **Aspect Ratio** | **Form Factor** | **Roundness** | **Convexity** |
|---|---|---|---|---|---|---|---|
| PS | 11,272 | 9.352 | 50.032 | 1.162 | 0.664 | 0.517 | 1.050 |
| PP | 7,956 | 2.877 | 6.470 | 1.092 | 0.698 | 0.551 | 1.049 |
| PET | 3,924 | 2.068 | 3.693 | 1.147 | 0.706 | 0.573 | 1.051 |
| PE | 614 | 3.175 | 8.313 | 1.146 | 0.730 | 0.578 | 1.040 |

From an image analysis perspective, these results indicate that the shape and size of MNPs are influenced by the type of bulk polymers they originated from. For instance, the significant differences in diameter and area suggest that the size distribution of MNPs can be used to distinguish between different polymer types. The differences in form factor and roundness indicate that the shape of the particles also varies by polymer type, which can affect their behavior in the environment. These findings can be used to develop more effective methods for identifying and quantifying MNPs in environmental samples. Moreover, the findings highlight the significant environmental impact of PS and PP due to their higher tendency to fragment into microscale, nanoscale, and larger particles. These small particles can penetrate various ecosystems, posing risks to aquatic life, and soil health, and can potentially enter the food chain. The lower fragmentation rates of PET and PE into smaller particles suggest these polymers may pose a lesser immediate risk in terms of MNP pollution. However, all four polymers contribute to the overall environmental burden of plastic pollution.

## 3.2. Experiment 1: MNP segmentation

Using the network configurations detailed in Table S2, we implemented multiple instance segmentation algorithms to evaluate MNP segmentation performance. Specifically, we tested YOLO-based segmentation models and Mask R-CNN with different backbone architectures to systematically assess segmentation quality.

According to these results in Table 3, YOLOv8 segmentation outperforms all other segmentation methods across all metrics. To investigate the shortcoming of existing algorithms, metrics were also reported based on particles size. Since images in our dataset have different scales, particle sizes are relative to the pixel size in images, not the real dimension of the particles.

Across all evaluated models, lower performance was most obvious for small particles (APs), which represent the primary source of segmentation error (Figure S3 and Section S3). This trend reflects the inherent challenges associated with small instances, including limited feature representation, reduced contrast, and increased sensitivity to localization inaccuracies. Notably, YOLO-based segmentation models demonstrated stronger performance on small and medium particles (APs and APm), whereas Mask R-CNN achieved superior results for large particles (APl), particularly in terms of mask generation accuracy.

**Table 3.** Comparison of evaluation metrics for segmentation of MNPs in the MiNa dataset. Lower or higher values being preferable are indicated by ↓ and ↑, respectively.

| Model | Backbone | AP50 ↑ | AP75 ↑ | AP ↑ |
|---|---|---|---|---|
| **Mask R-CNN** | ResNet 101 | 75.878 | 44.723 | 43.541 |
| | ResNext 101 | 71.061 | 45.145 | 42.365 |
| **YOLO Segmentation** | yolov8n-seg | **81.58** | **53.03** | **49.72** |
| | yolov26n-seg | 73.06 | 44.46 | 42.59 |

Qualitative evaluation of the segmented particles indicates that oversegmentation and fragmented masks constitute major limiting factors affecting the performance metrics across all methods. This issue becomes more pronounced for larger particles and in densely crowded regions. Among the evaluated models, Mask R-CNN with the ResNeXt backbone demonstrates greater robustness to these artifacts, as illustrated in Figure 3.

### 3.3. Experiment 2: MNP identification

This experiment investigates the possibility of correlation between polymer type and morphological conditions of degraded polymer under the same degradation conditions. For this purpose, particles have been labeled according to their polymer type. In this experiment, we have used object classification networks that predict the bounding box and class of the test samples. We call this experiment identification since it can predict the location of the particles, count them, and predict the polymer type. Table 4 summarizes the detection performance metrics. Faster R-CNN achieves the highest recall, whereas RT-DETR demonstrates superior precision.

**Table 4.** Performance comparison of object detection models for polymer particle identification. The table reports mAP50, precision (P), recall (R), and F1-score for RT-DETR, Faster R-CNN (ResNet-101 and ResNeXt-101 backbones), and YOLO (yolov26s and yolov8s). YOLOv26s achieves the highest mAP50 and F1-score, while Faster R-CNN attains the highest recall. RT-DETR demonstrates competitive precision with a larger backbone. Lower or higher values being preferable are indicated by ↓ and ↑, respectively.

| Model | Backbone | mAP50 ↑ | Precision (P) ↑ | Recall (R) ↑ | F1 ↑ |
|---|---|---|---|---|---|
| RT-DETR | Large | 74.00 | **74.70** | 70.10 | 72.33 |
| Faster R-CNN | ResNet 101 | 54.49 | 58.00 | 75.10 | 65.45 |
| | ResNext 101 | 57.12 | 56.60 | **75.70** | 64.77 |
| YOLO | yolov26s | **77.30** | 74.40 | 72.30 | **73.33** |
| | yolov8s | 74.20 | 72.80 | 69.30 | 71.01 |

Overall, YOLOv26 outperforms Faster R-CNN in terms of AP50, despite using a smaller backbone. Faster R-CNN, like every R-CNN-based detector, is highly sensitive to anchor size configuration. The anchor settings reported in Table S2 correspond to those yielding the highest AP50. Although precision can be further improved through anchor recalibration, this typically results in a reduction in recall, reflecting the inherent precision-recall trade-off.

Figure 4 illustrates the detection of MNPs and the prediction of their corresponding polymer types. Since each sample contains only a single polymer type, all particles within a given patch share the same ground-truth label. Qualitative assessment indicates that smaller particles and lower spatial scales are more prone to misclassification (Figure 4a, c).

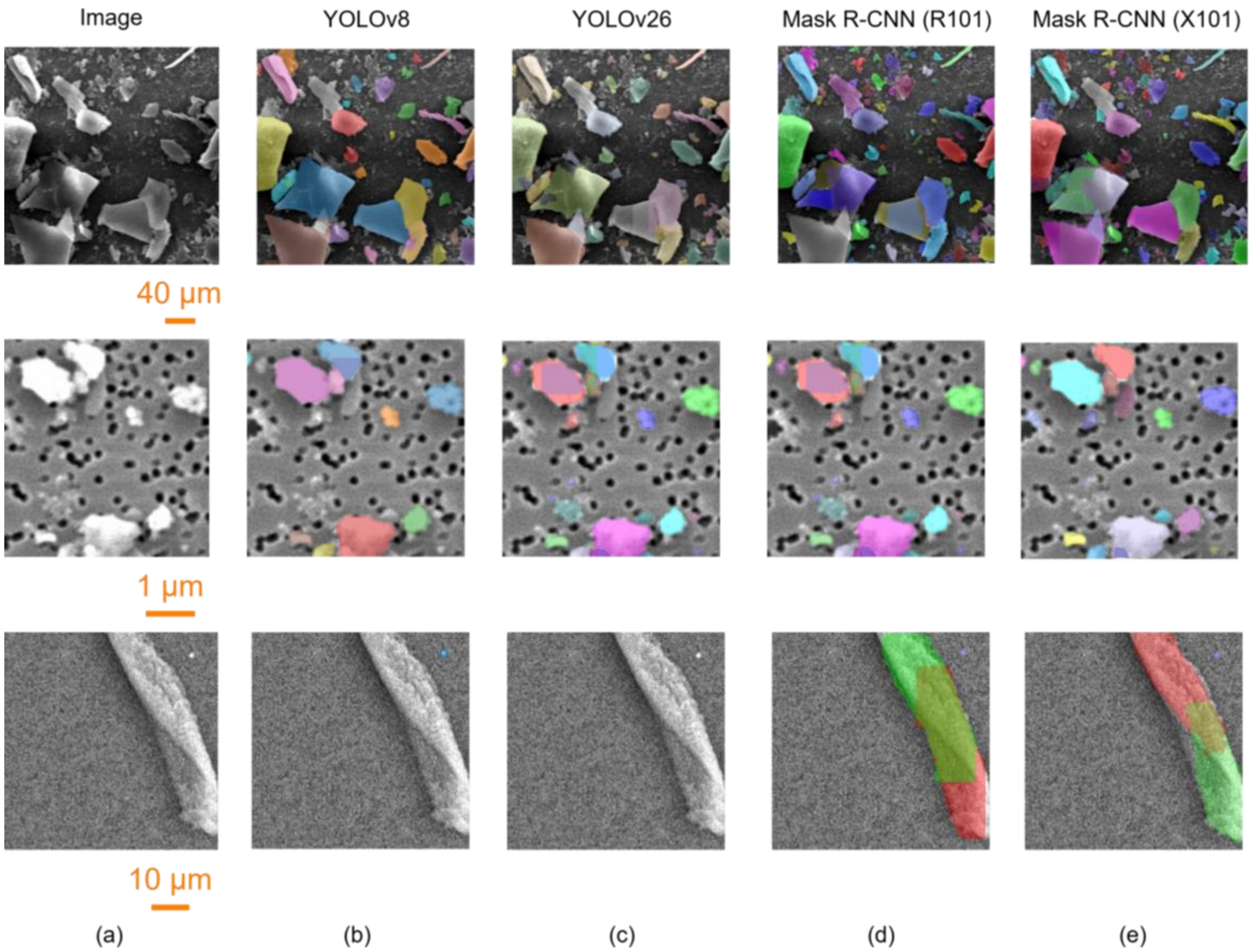

**Figure 3.** Qualitative comparison of particle segmentation results obtained using YOLO-based and Mask R-CNN models. (a) Representative image patches selected from the test set on which inference was performed. (b, c) Segmentation outputs from YOLOv8-seg and YOLOv26-seg (small variants). While YOLO models perform adequately for small particles, they exhibit limitations in accurately capturing the boundaries of medium and large particles. Oversegmentation and fragmented masks are frequently observed, particularly in YOLOv26, contributing to reduced performance for larger particle sizes. (d, e) Segmentation results from Mask R-CNN models using ResNet-101 (R101) and ResNeXt-101 (X101) backbones. Mask R-CNN provides more consistent boundary definition for large particles. However, oversegmentation persists in densely populated regions. The ResNeXt-101 backbone demonstrates improved segmentation consistency across particle size ranges compared to ResNet-101.

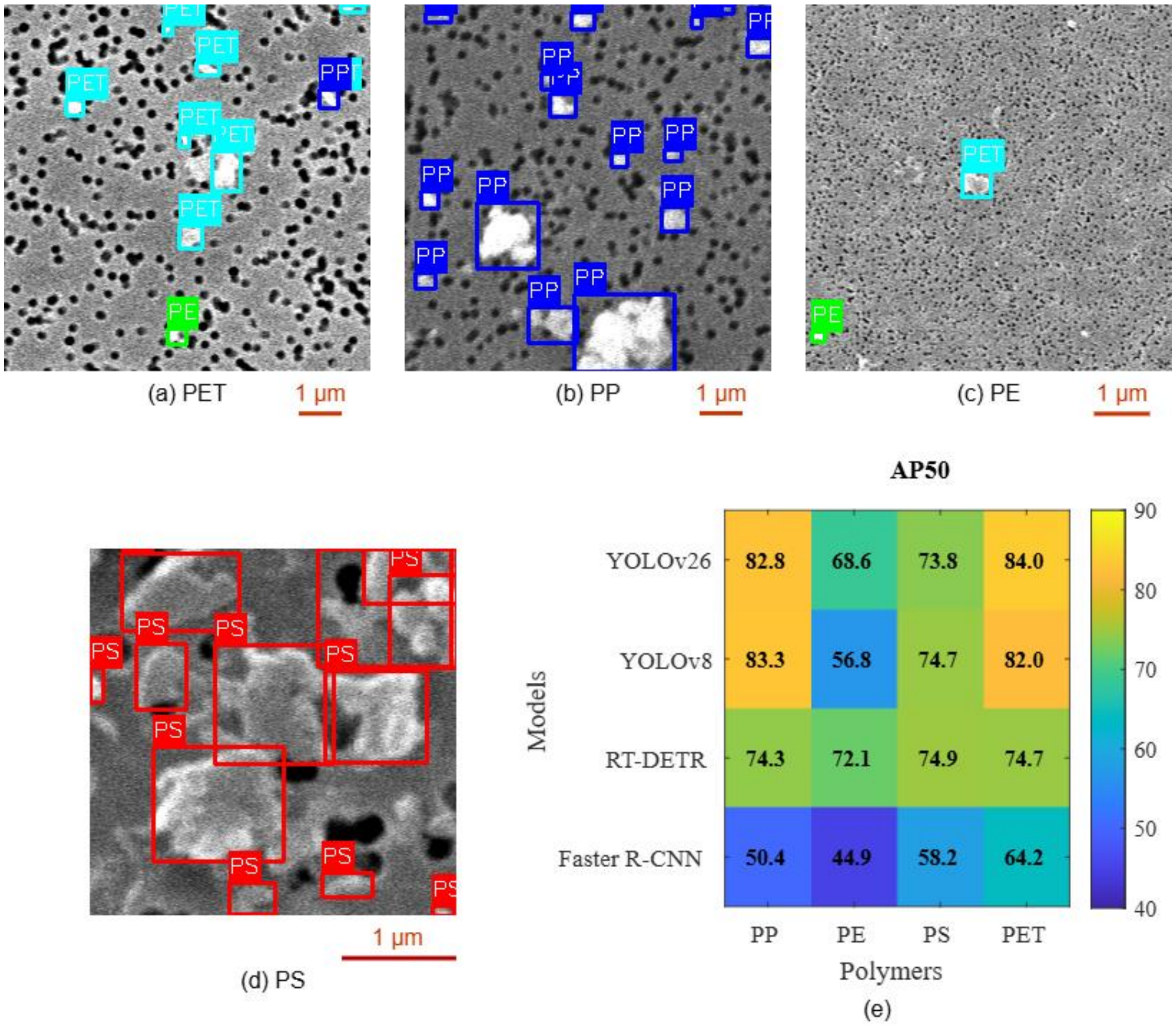

**Figure 4.** Detection and classification of polymer types using YOLOv26. Most misclassifications occur for smaller particles and at smaller spatial scales (a-b) PET and PP particles, despite their smaller average area, show consistent detection and classification performance across models. (c) PE exhibits the lowest particle count among the four polymers, which may contribute to its comparatively lower performance due to the lower training samples. (d) PS has the highest number of particles in the images which makes more overlapping scenarios in the images. (e) The results suggest that, under consistent degradation conditions, morphological features captured in SEM images are sufficiently discriminative to correlate with polymer type. Although YOLOv26 has better overall performance, RT-DETR shows a uniformly acceptable performance across all polymer classes. Across all evaluated networks, degraded particles exhibit distinguishable morphological features that enable polymer-type discrimination. RT-DETR demonstrates the most stable AP50 performance across all polymer categories.

In contrast, at larger spatial scales, where morphological features are more distinct, the networks demonstrate improved predictive performance. Notably, YOLOv26 achieves an AP50 of 77.3% for polymer-type classification (Figure 4).

Figure 4d presents the per-class breakdown of polymer identification performance based on morphological features observed in SEM images. Across all architectures, PET and PP consistently achieve higher AP50 values, whereas PE remains the most challenging class. This consistent ranking across YOLO, RT-DETR, and Faster R-CNN suggests that classification performance is primarily governed by the intrinsic morphological separability of degraded polymers rather than the detector architecture itself. In particular, PE exhibits lower precision–recall balance and reduced AP, indicating stronger morphological overlap with other polymer types. Conversely, PET and PP demonstrate more distinctive surface textures and structural patterns that facilitate reliable discrimination. These results support the hypothesis that meaningful morphology-polymer correlations exist in SEM imagery, while also highlighting inherent limits imposed by inter-class morphological similarity.

### 3.4. Experiment 3: Evaluation of Foundation Models

While transformers excel in large-scale domains, their application to MNPs is limited by the scarcity of labeled SEM datasets[41]. We evaluated the transferability of SAM[36] and micro-SAM[30] using Automatic Instance Segmentation (AIS) and Automatic Mask Generation (AMG). Although AMG provided higher surface accuracy, both zero-shot methods were outperformed by prompting SAM with bounding boxes from our fine-tuned RT-DETR model (see Table S3).

Qualitative analysis revealed that while foundation models segment well-contrasted particles, they lack the semantic specificity to distinguish MNPs from background pores or morphological noise without expert prompting (Figure 5). By using RT-DETR as a deterministic prompting engine, we eliminated human bias and significantly reduced inference time on an NVIDIA RTX 4080 GPU. Cross-dataset validation on fragmented microplastics[12] yielded a mean IoU of 76.22% and a MAPE of 14.56% without additional training (Figure S5). The integration of the MiNa dataset with our RT-DETR-prompted SAM architecture establishes a robust, objective framework for the rapid segmentation of complex SEM imagery, providing the field with the essential tools to automate high-throughput MNP quantification across diverse environmental matrices.

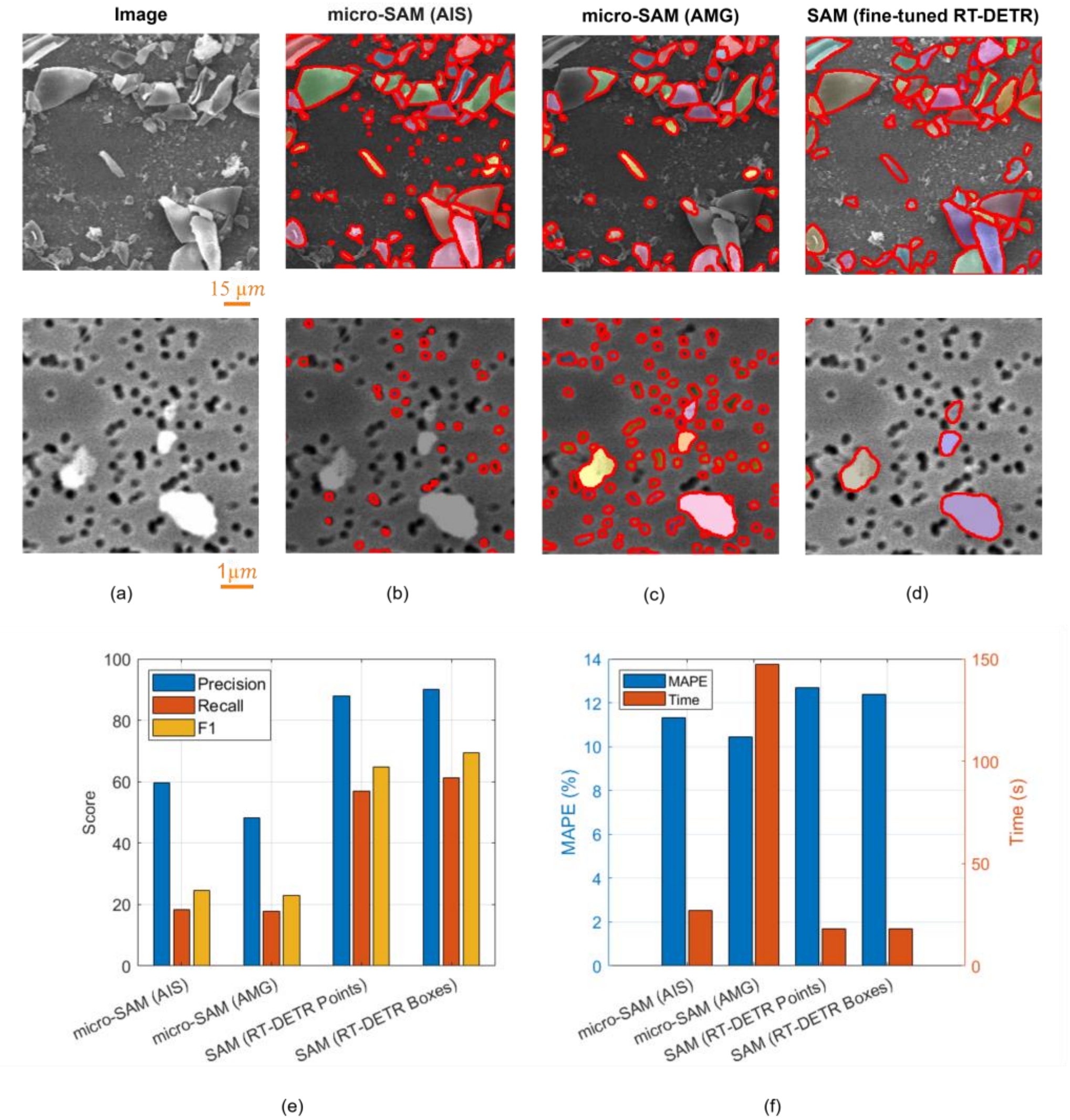

**Figure 5.** Zero-shot segmentation results using SAM-based foundation models on representative SEM micrographs from the MiNa dataset. (a) Representative SEM micrographs from the MiNa dataset. (b) With AIS prompting, particles are accurately segmented in relatively uniform backgrounds; however, performance degrades in the presence of porous structures. (c) AMG improves robustness to such backgrounds and detects particles within pores, but often over-segments by including pore regions and fails to resolve overlapping particles. In contrast, the fine-tuned RT-DETR model provides precise point and box prompts to SAM, leading to significantly improved segmentation quality and particle delineation (d). These results underscore the critical importance of accurate prompt generation when leveraging general-purpose foundation models for automated segmentation and quantitative analysis.

## 4. Conclusion

The analysis of MNPs emerged from consumer plastics over 12 weeks under environmentally relevant conditions provides valuable insights into less explored areas of MNPs. Quantitative analysis highlighted the potential impact of PP and PS food containers, which released thousands of MNPs to the environment in a short span. Additionally, commercial PET water bottles produced a considerable number of nanoplastics, which can pose significant health risks. In contrast, PE plastic bags, which are being banned in many jurisdictions, released the least number of MNPs.

Our goal in developing the MiNa dataset was to create an open-source and diverse SEM micrograph–based dataset of MNPs originating from different polymer types. To the best of our knowledge, this dataset is the first to annotate individual particles according to their polymer type, enabling the investigation of morphological patterns associated with different polymers in realistic environmental conditions. Visual inspection of the particles suggests that distinct physical and morphological characteristics are present across polymer types, indicating that these patterns can be used to classify particles into different polymer classes with reasonable accuracy.

The MiNa dataset provides the first open-source SEM benchmark for MNPs annotated by polymer type, enabling systematic study of morphology-chemistry relationships under realistic degradation. Benchmarking confirms that deep learning supports automated MNP quantification, but reveals persistent challenges in crowded scenes and broad size distributions. This work establishes a foundation for high-throughput AI monitoring, offering a pathway toward zero-shot segmentation of complex micrographs. Future efforts should expand into mixed-particle datasets and real-world environmental samples containing biological matter to enhance model generalization and practical utility for global plastic pollution monitoring.

## Authorship Contribution Statement

**Hadi Rezvani**: Conceptualization, Methodology, Data Curation, Data Annotation, Writing and Visualization, Analysis; **Navid Zarrabi**: Data Annotation, Benchmarking, Software, Methodology, Analysis, Writing and Visualization; **Ishaan Mehta**: Data Annotation and Review, Data Preprocessing; **Christopher Kolios**: Data Preprocessing, Annotation; **Hussein Ali Jaafar**: Annotation; **Cheng-Hao Kao**: Annotation; **Sajad Saeedi**: Writing – Review & Editing,

Supervision, Methodology, Funding Acquisition, Conceptualization; **Nariman Yousefi**: Writing – Review & Editing, Supervision, Methodology, Funding Acquisition, Conceptualization.

**Funding**

This work was supported by the Natural Sciences and Engineering Research Council of Canada's (NSERC) Discovery Grants program (grant no. RGPIN-2020-05436 and RGPIN-2020-04659), Canada Foundation for Innovation's John R. Evans Leaders Fund (grant no. 39659), Environment and Climate Change Canada's (ECCC) Increasing Knowledge on Plastic Pollution Initiative (contract no. GCXE21S057), and Social Sciences and Humanities Research Council of Canada's (SSHRC) New Frontiers in Research Fund-Exploration program (grant no. NFRFE-2022-00746).

**Abbreviations**

**MNPs** micro- and nanoplastics

**SEM** Scanning Electron Microscopy

**IoU** Intersection over Union

**PE** Polyethylene

**PS** Polystyrene

**PP** Polypropylene

**PET** Polyethylene Terephthalate

**R-CNN** Region-based Convolutional Neural Network

**ML** Machine Learning

**CNN** Convolutional Neural Network

**MiNa** Micro- and Nanoplastics dataset